\documentclass[10pt,twocolumn,letterpaper]{article}

\usepackage{cvpr}              % To produce the CAMERA-READY version
\usepackage{graphicx}
\usepackage{amsmath}
\usepackage{amssymb}
\usepackage{booktabs}

\usepackage[pagebackref,breaklinks,colorlinks]{hyperref}

\usepackage[capitalize]{cleveref}
\crefname{section}{Sec.}{Secs.}
\Crefname{section}{Section}{Sections}
\Crefname{table}{Table}{Tables}
\crefname{table}{Tab.}{Tabs.}

\def\cvprPaperID{2200} % *** Enter the CVPR Paper ID here
\def\confName{CVPR}
\def\confYear{2023}

\usepackage[inline]{enumitem}

\usepackage{xcolor}
\usepackage{skak}
\newcommand\blfootnote[1]{%
  \begingroup
  \renewcommand\thefootnote{}\footnote{#1}%
  \addtocounter{footnote}{-1}%
  \endgroup
}

\newcommand{\Section}[1]{\cref{sec:#1}}
\newcommand{\Figure}[1]{\cref{fig:#1}}
\newcommand{\Table}[1]{\cref{tab:#1}}
\newcommand{\eq}[1]{Eq.~\eqref{eq:#1}}

\newcommand{\ill}{\mathbf{l}}
\newcommand{\feat}{\mathbf{f}}
\newcommand{\pose}{\mathbf{p}}
\newcommand{\triplaneG}{\mathcal{G}_\text{tri}}
\newcommand{\upsampleU}{\mathcal{U}}
\newcommand{\ray}{\mathbf{r}}
\newcommand{\view}{\mathbf{d}}
\newcommand{\bc}{\mathbf{c}}
\newcommand{\density}{\sigma}
\newcommand{\bw}{\mathbf{w}}
\newcommand{\bx}{\mathbf{x}}
\newcommand{\real}{\mathbb{R}}
\newcommand{\bI}{\mathbf{I}}
\newcommand{\bd}{\mathbf{d}}
\newcommand{\normal}{\mathbf{n}}
\newcommand{\reflect}{\boldsymbol{\omega}_r}

\newcommand{\shiny}{{k}_s}
\newcommand{\intomega}{\boldsymbol{\omega}}

\newcommand{\ours}{FaceLit}

\renewcommand{\paragraph}[1]{\vspace{0.5em}\noindent\textbf{#1}}

\begin{document}

\title{\ours: Neural 3D Relightable Faces}

\author{Anurag Ranjan$^\symknight$ \; \; \;
Kwang Moo Yi$^{\symknight\symrook}$ \; \; \;
Jen-Hao Rick Chang$^\symknight$ \; \; \;
Oncel Tuzel$^\symknight$ \\
$^\symknight$Apple \; \; \;
$^\symrook$The University of British Columbia
}

\twocolumn[{%
\renewcommand\twocolumn[1][]{#1}%
\maketitle
\begin{center}
    \newcommand{\teaserwidth}{\textwidth}
\vspace{-0.3cm}
   \includegraphics[width=0.96\linewidth]{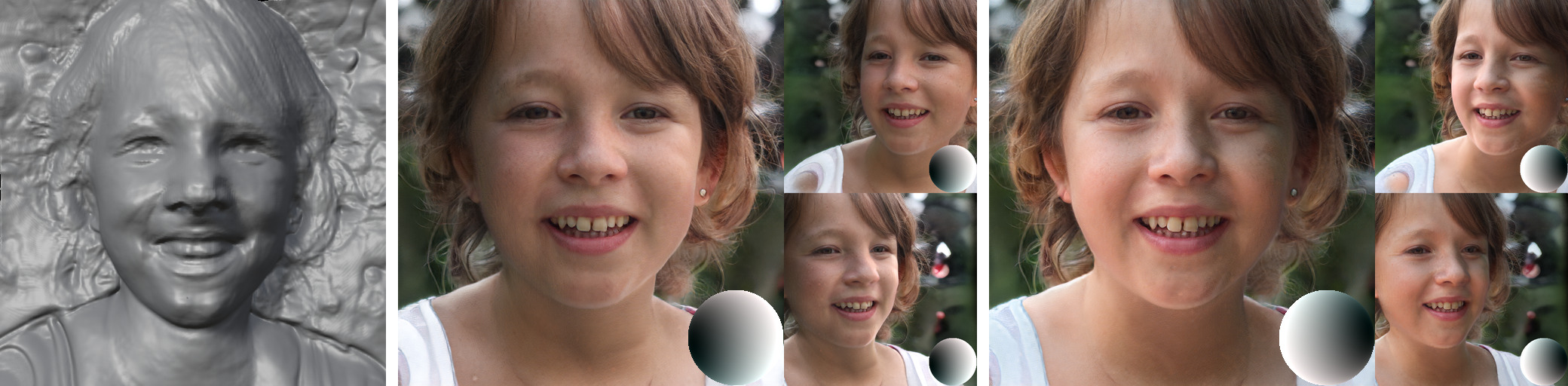} \\
  \includegraphics[width=0.96\linewidth]{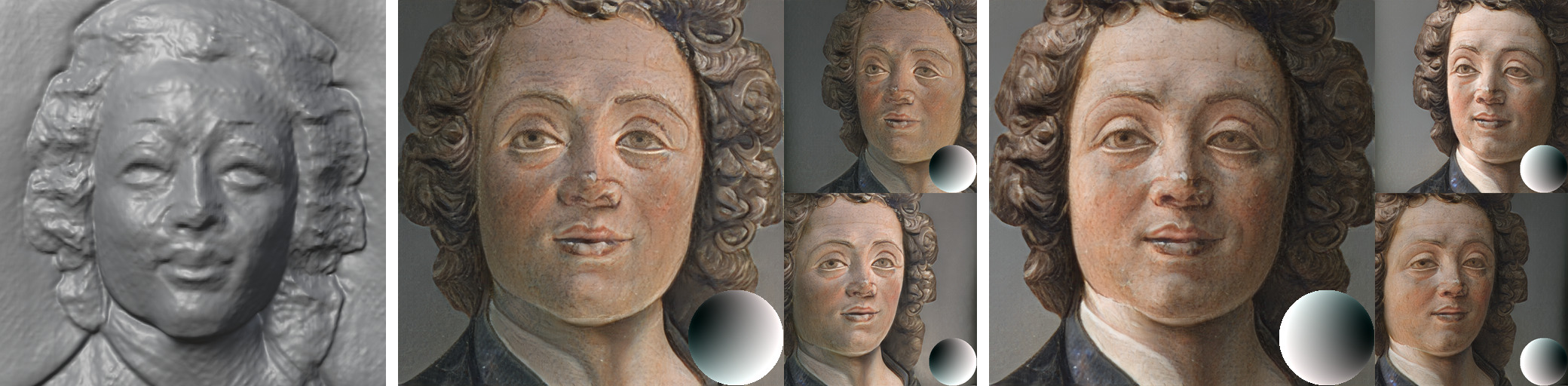}
%   }
   \captionof{figure}{Generated samples from our model. \textbf{Left}: 3D reconstruction visualization. \textbf{Center and Right}: Rendered faces using 2 different illumination conditions under 3 different poses. Illumination visualization using spherical harmonics~\cite{ramamoorthi2001efficient}.
}
\label{fig:teaser}
\end{center}%
}]

\maketitle

%%%%%%%%% ABSTRACT
\begin{abstract}
\vspace{-0.2cm}
{
We propose a generative framework, \ours, capable of generating a 3D face that can be rendered at various user-defined lighting conditions and views, learned purely from 2D images in-the-wild without any manual annotation. Unlike existing works that require careful capture setup or human labor, we rely on off-the-shelf pose and illumination estimators. With these estimates, we incorporate the Phong reflectance model in the neural volume rendering framework. Our model learns to generate shape and material properties of a face such that, when rendered according to the natural statistics of pose and illumination, produces photorealistic face images with multiview 3D and illumination consistency. Our method enables photorealistic generation of faces with explicit illumination and
view controls on multiple datasets – FFHQ, MetFaces and CelebA-HQ. We show state-of-the-art photorealism among 3D aware GANs on FFHQ dataset achieving an FID score of 3.5.} \blfootnote{corresponding authors: \texttt{\{anuragr, otuzel\}@apple.com}. \\ $^\symrook$ work done while at Apple.}
\end{abstract}
\vspace{-0.5cm}

%%%%%%%%% BODY TEXT
\section{Introduction}
\label{sec:intro}
Learning a 3D generative model from 2D images has recently drawn much interest~\cite{sitzmann2019srn, niemeyer2021giraffe, gu2021stylenerf}.
{%
Since the introduction of Neural Radiance Fields (NeRF)~\cite{mildenhall2020nerf}, the quality of images rendered from a 3D model~\cite{gu2021stylenerf, Chan2022eg3d} has improved drastically, becoming as photorealistic as those rendered by a 2D model~\cite{karras2020stylegan2}.
}%
While some of them~\cite{chan2021pigan, rebain2022lolnerf} rely {purely} on 3D representations to {deliver}
3D consistency and pay the price of {decreased} photorealism, more recent work~\cite{Chan2022eg3d} has further shown {that this can be avoided and extreme photorealism can be obtained through a hybrid setup}.
However, {even so,} a shortcoming of these models is that the components that constitute the scene---the geometry, appearance, and the lighting---are all \emph{entangled} and are thus not controllable using user defined inputs.

Methods have been proposed to break this entanglement~\cite{boss2021neural, boss2021nerd, zhang2021nerfactor},
however, they require multiview image collections of the scene being modeled and are thus inapplicable to images in-the-wild where such constraint cannot be satisfied easily.
Boss et al.~\cite{boss2022samurai} loosens this constraint to images of different scenes, but they still require the same object to be seen from multiple views at the end.
Moreover, these methods are not generative and therefore need to be trained for each object and cannot generate new objects.
For generative methods~\cite{gu2021stylenerf, Chan2022eg3d, chan2021pigan}, geometry and illumination remains entangled.
In this work, we demonstrate that {one does not require multiple views, and} the variability and the volume of already existing datasets~\cite{karras2019stylegan, karras2018progressive, karras2020training} are enough to learn a disentangled 3D generative model.

We propose \emph{\ours}, a framework that learns a \emph{disentangled} 3D model of a face, purely from images; see~\Figure{teaser}.
The high-level idea behind our method is to build a rendering pipeline that is \emph{forced} to respect physical lighting models~\cite{ramamoorthi2001efficient, phong1975illumination}, similar to \cite{zhang2021nerfactor} but in a framework friendly to 3D generative modeling, and one that can leverage off-the-shelf lighting and pose estimators~\cite{feng2021deca}.
In more detail, we embed the {physics-based} illumination model using Spherical Harmonics~\cite{ramamoorthi2001efficient} within the recent generative Neural Volume Rendering pipeline, EG3D~\cite{Chan2022eg3d}.
We then simply train for realism, and since the framework has to then obey physics to generate realistic images, it naturally learns a disentangled 3D generative model.

Importantly, the way we embed physics-based rendering into neural volume rendering is the core enabler of our method.
As mentioned, to allow easy use of existing off-the-shelf illumination estimators~\cite{feng2021deca}, we base our method on Spherical Harmonics.
We then model the diffuse and specular components of the scene via the Spherical Harmonic coefficients associated with the surface normals and the reflectance vectors, where the {diffuse reflectance}, the normal vectors and the material {specular reflectance} are generated by a neural network. 
While simple, our setup allows for effective disentanglement of illumination from the rendering process.

We show the effectiveness of our method using three datasets FFHQ~\cite{karras2019stylegan}, CelebA-HQ~\cite{karras2018progressive} and MetFaces~\cite{karras2020training} and obtain state-of-the-art FID scores among 3D aware generative models.
Furthermore, to the best of our knowledge, our method is the very first generative method that can generate 3D faces with controllable scene lighting. {Our code is available for research purposes at \url{https://github.com/apple/ml-facelit/}.}

To summarize, our contributions are:
\vspace{-0.7em}
\begin{itemize}[leftmargin=*]
\setlength\itemsep{-.3em}
    \item we propose a novel framework that can learn a disentangled 3D generative model of faces from single views, with which we can render the face with different views and under various lighting conditions;
    \item we introduce how to embed an illumination model in the rendering framework that models the effects of diffuse and specular reflection.
    \item we show that our method can be trained \emph{without} any manual label and simply with 2D images and an off-the-shelf pose/illumination estimation method.
    \item we achieve state-of-the-art FID score of 3.5 among 3D GANs on the FFHQ dataset improving the recent work~\cite{Chan2022eg3d} by 25\%, relatively. 
\end{itemize}

\section{Related work}
\label{sec:related}

\begin{figure*}[t]
    \centering
    \includegraphics[width=\linewidth]{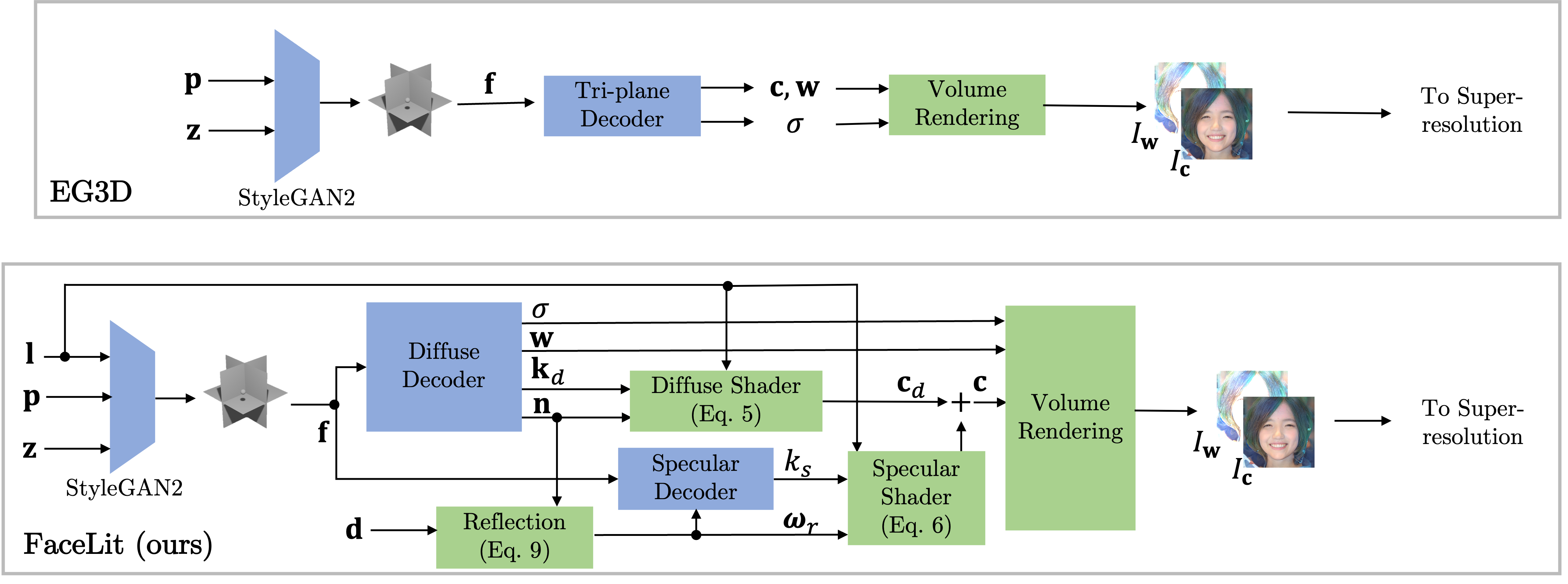}
    \caption{\textbf{Overview:} Our generation process samples a 3D face from a latent $\bf z$ conditioned on pose $\bf p$ and illumination $\ill$. The generator uses a StyleGAN2 backbone with a tri-plane feature representation $\bf f$ introduced by EG3D~\cite{Chan2022eg3d} and introduces illumination modeling using diffuse and specular decoders. Green modules are non-trainable fixed differentiable functions. See \S \ref{sec:generator} for details.}
    \label{fig:pipeline}
\end{figure*}

We first discuss works that focus on decomposing a scene into shape, appearance, and lighting, then discuss those that focus on 3D generative models.

\paragraph{Decomposing a scene into shape, appearance and lighting.}
Neural Radiance Fields (NeRF)~\cite{mildenhall2020nerf} learn a {5D radiance field}
where the aspects of the scene such as shape, appearance and lighting are jointly modeled by a neural network.
{While effective, this results in an entangled representation of the scene where only the camera pose can be controlled.
In fact, besides controllability, in Ref-NeRF~\cite{verbin2022refnerf}, it was demonstrated that explicitly allowing diffuse and specular rendering paths within the deep architecture, thus having in a sense architectural disentanglement leads to improved rendering quality.
Thus, it is unsurprising that various methods have been proposed for disentanglement.}

Recent works~{\cite{boss2021nerd, zhang2021nerfactor} use}
a Bidirectional Reflectance Distribution function (BRDF) to learn a scene representation using multiview images. 
Neural-PIL~\cite{boss2021neural} computes a pre-integrated lighting map from multiview image collections and models diffuse and specular reflectance in the scene.
SAMURAI~\cite{boss2022samurai} operates on image collections of an object under different views and lighting conditions {with various backgrounds thus reducing the strictness of the multiview constraint.}
{%
However, even with reduced constraint, they still require the \emph{same} object of interest to be in the scene from multiple views. 
In our case, we are interested in the problem setting when only single views are available, which none of these methods can be trivially extended to.
}%

\paragraph{3D generative models.}
Generative Adversarial Networks (GAN)~\cite{goodfellow2020generative} trained from single-view images have been shown to be successful in  generating photorealistic face images~\cite{karras2019stylegan, karras2020stylegan2}.
Early works to enforce 3D consistency on generated images relied on conditioning the GAN with pose~\cite{nguyen2019hologan} and other attributes~\cite{ghosh2020gif, ruiz2020morphgan} such as expression and lighting.
However, these methods do not model the physical rendering process like NeRFs.
This leads to inconsistent 3D geometry---rotating images generated from these 2D-based models result in change of shape and appearance and not just the view point.
Follow up work~\cite{chan2021pigan, rebain2022lolnerf, gu2021stylenerf, zhao2022gmpi, schwarz2022voxgraf} thus uses volume rendering---which brings 3D consistency by construction---on top of GANs to force 3D consistent representation. {Furthermore, recent work~\cite{tan2022volux, pan2021shadegan, prao2022vorf} use illumination modeling on top of volume rendering.}
While this allowed the models to be 3D consistent, their generated image quality is not as photorealistic as their 2D counterparts~\cite{karras2020stylegan2}.
{%
EG3D~\cite{Chan2022eg3d} thus proposes a tri-plane representation, and a hybrid framework that achieves the level of photorealism similar to 2D GAN frameworks.
}%
{%
In a different direction, Cao et al.~\cite{Cao22} utilize a massive dataset of multiple views of human faces to build a generative model that are conditioned on facial expressions, that are then adapted to a subject of interest for a controllable 3D model.
}%

{%
Regardless of whether these models aim for unconditional generation or controllability, they, however, do not disentangle geometry from illumination and thus cannot be relighted, limiting their application towards a fully controllable generative 3D model.
}%

\subsection{Preliminaries: The EG3D framework}

As we base our framework on the EG3D framework~\cite{Chan2022eg3d}, we first briefly explain the pipeline in more detail before discussing our method.
The core of the EG3D pipeline is the use of tri-plane features, which allows the use of well-studied 2D CNNs for generating deep features to be used for volume rendering.

As shown in \Figure{pipeline}, EG3D uses a tri-plane generator $\triplaneG$ with a StyleGAN2~\cite{karras2020stylegan2} backbone conditioned on camera pose $\pose$ to generate feature maps.
These feature maps are rearranged to obtain tri-plane features $\feat^{XY}, \feat^{YZ}, \feat^{XZ}$ along 3 orthogonal planes.
A decoder neural network is then used to regress color $\bc$ and density $\density$ and additional features $\bw$ at a given location $\bx \in \real^3$ from the tri-plane features.
A color image $\bI_\bc$ is then obtained by aggregating the values $\bc, \sigma$ by volume rendering along the ray $\ray$ given by
\begin{equation}
\bI_\bc (\ray) = \int_{t_n}^{t_f} T(t) \density(\ray(t)) \bc(\ray(t), \bd) dt
,
\label{eq:volume_rendering}
\end{equation}
where spatial locations are sampled within the near and far plane locations as $t \in [t_n, t_f]$, $\bd$ is the viewing direction, and the transmittance 
\begin{equation}
    T(t) = \exp \left ( - \int_{t_n}^t \density(\ray(u)) du \right )
    .
\end{equation}
{This volume rendering of $\bI_\bc$ is performed at a relatively low resolution to be memory efficient, and an upsampling is performed for higher resolution images.
Hence, an additional} $n_w$-channel feature image $\bI_\bw$ is rendered by tracing over the features $\bw \in \mathbb{R}^{n_w}$, {which is then used for generating} the final image $\bI^+_\bc = \upsampleU(\bI_\bc, \bI_\bw)$, with the super-resolution network $\upsampleU$.
{%
The framework is then trained to make $\bI^+_\bc$ as realistic as possible through GAN training setup, with the discriminator being conditioned on the poses $\pose$.
}%

{%
Note here that, as the discriminator is conditioned on the pose, the pose must be provided for each training image; which has been shown to be effective in delivering better 3D consistency~\cite{Chan2022eg3d}.
In the case of EG3D~\cite{Chan2022eg3d}, these poses are obtained using Deng et al.~\cite{deng2019facerecon}.
Similarly, we will rely on DECA~\cite{feng2021deca} since it provides the estimates of poses, as well as illumination.
}%

\section{Method}
\label{sec:method}

While modeling a scene with a radiance field is effective in rendering from novel viewpoints, it hinders our capability to relight the scene with different illumination conditions {as illumination is entangled with the appearance and shape}. 
To relight with unseen illumination conditions, we incorporate physics-based shading into the forming of radiance fields, {thus disentangling it by construction}.

In the following subsections we first discuss the illumination model that we propose which allows us to achieve disentanglement, then detail how we implement the model into a deep generative model.

\subsection{Illumination model}
\label{sec:illumination}

To explicitly constrain the rendering process on illumination we use a simplified version of the Phong reflectance model~\cite{phong1975illumination}.
The color at a location is computed using
{%
\begin{equation}
    \bc = \int_{\intomega} \left(\mathbf{k}_d \odot \left(\normal \cdot {\intomega} \right) L^d({\intomega}) + k_s \left(\reflect \cdot {\intomega} \right)^\alpha L^s(\intomega)\right) \,d{\intomega}
    ,
    \label{eq:phong}
\end{equation}
}%
where $\normal$ is its normal, $\reflect$ is the reflection direction given by Eq.~(\ref{eq:reflection_direction}),  $\mathbf{k}_d \in  \mathbb{R}^3$ is the diffuse reflectance,  $k_s  \in \mathbb{R}$
is the specular reflectance coefficient, $\alpha$ is the shininess constant, and $L^d$ and $L^s: \mathbb{R}^3 \rightarrow \mathbb{R}^3$ are the diffuse and specular environment maps (distance light distributions) respectively parameterized by incident light direction, $\intomega$ on the surface of the unit sphere. The operator  $\odot$ is element-wise multiplication and $\cdot$ is the dot product. For brevity, when we element-wise multiply a scalar with a vector, we assume each element of the vector is multiplied by the scalar. 
Here, the first term computes the diffuse color $\bc_d$ and the second term computes the specular color $\bc_s$.

\paragraph{Further simplification via Spherical Harmonics.}
We assume a single environment map for diffuse and specular, $L^d = L^s = L$. To speed up rendering Eq.~(\ref{eq:phong}) and efficient representation, we follow Ramamoorthi et al.~\cite{ramamoorthi2001efficient}, and pre-integrate the environment map to compute the irradiance environment map 
\begin{equation}
    E(\normal) = \int_{\intomega}  \left(\normal \cdot {\intomega} \right) L({\intomega}) \,d{\intomega}.
    \label{eq:irradiance}
\end{equation}
The irradiance environment map can be efficiently (approximately) represented in Spherical Harmonics (SH) basis using only 9 basis functions. See~\cite{ramamoorthi2001efficient} for details. 

Consider that irradiance environment map is represented by SH coefficients $\mathbf{l}_k \in \mathbb{R}^3$ with SH basis $H_k : \mathbb{R}^3 \rightarrow \mathbb{R}$ and $k \in [1,9]$.
We can fold all illumination-related terms in~\eq{phong} using the SH basis functions.
Thus, the diffuse term can be rewritten as 
\begin{equation}
    \mathbf{c}_d = \mathbf{k}_d \odot \sum_k \ill_k  H_k(\normal)
    .
    \label{eq:diffuse_shading}
\end{equation}

For the specular component we assume that $\alpha = 1$\footnote{Using $\alpha = 1$ allows us to use the same irradiance environment map to compute the specular color.} in~\eq{phong} which, with the SH basis again folding in the illumination terms gives us 

\begin{equation}
    \bc_s =  k_s \sum_k   \ill_k  H_k(\reflect)
    .
    \label{eq:specular}
\end{equation}
{%
Note here that unlike in \eq{diffuse_shading} we use $\reflect$ to retrieve the irradiance environment map values  
rather than $\normal$, as we are interested in the specular component, which reflects off the surface.
}%
The final color is then a composition of specular and diffuse components 
\begin{equation}
    \bc = \bc_d + \bc_s
\end{equation}

{%
As we will show in \Section{experiments}, this simple formulation works surprisingly well, with the illumination being explicitly factored out.
In other words, by controlling $\ill_k$ in~\eq{diffuse_shading} and \eq{specular}, one can control how the face renders under different illuminations. Although, we do not model other effects of light on the skin, such as subsurface scattering~\cite{krishnaswamy2004sss}, we expect our model to account for it from the training process.

}%

\subsection{Generator}
\label{sec:generator}

To imbue a 3D generative model with explicitly controllable illumination, 
we condition the tri-plane generator on both the camera pose $\pose$ and the illumination $\ill$.
{%
This allows us to take into account the distribution of illumination conditions within our training dataset, similar to how pose was considered in EG3D~\cite{Chan2022eg3d}.
}%
Mathematically, we write our tri-plane generator as 
\begin{equation}
\feat^{XY}, \feat^{YZ}, \feat^{XZ} = \triplaneG \left( \pose, \ill \right)
   . 
\end{equation}
For a given point in space $\bx\in \mathbb{R}^3$, the aggregated features are obtained using $\feat_\bx = \feat^{XY}_\bx + \feat^{YZ}_\bx + \feat^{XZ}_\bx$. 
However, as discussed earlier, simply conditioning the tri-plane generator $\triplaneG$ alone is not enough to enable explicit and consistent control over the camera pose and illumination---there is no guarantee that the generated content will remain constant while illumination and camera pose changes.
{%
We thus utilize $\feat_\bx$ and apply the illumination model in \Section{illumination}.
}%

{%
Specifically, as shown in \Figure{pipeline}, instead of directly regressing the color $\bc$ and the density $\density$ at a given point as in EG3D~\cite{Chan2022eg3d}, we decode $\feat_\bx$ using diffuse and specular decoders. We then apply shading through \eq{diffuse_shading} and \eq{specular} to obtain the diffuse color $\mathbf{c}_d$ and the specular color $\mathbf{c}_s$ of a point.
}%

\paragraph{Diffuse decoder.} % -- \eq{diffuse_shading}.} 
We regress the diffuse reflectance $\mathbf{k}_d$, normal $\normal$ and density $\density$ at a point given tri-plane features $\mathbf{f_x}$ using diffuse decoder (see \Figure{pipeline}).
We then apply \eq{diffuse_shading} to obtain the diffuse color $\bc_d$.
{Here, as in other NeRF work~\cite{zhang2021nerfactor,verbin2022refnerf} we opt to directly regress the normals, as we also found that using the derivative of the density to be unreliable for training.}

\paragraph{Specular decoder.} % -- \eq{specular}.} 
We regress the specular reflectance coefficient $\shiny$ using the specular decoder.
We also compute the reflection direction which is a function of the view direction $\view$ and the normal $\normal$ given by
\begin{equation}
    \reflect = \view - 2 (\view \cdot \normal)\normal
    .
    \label{eq:reflection_direction}
\end{equation}
We then obtain the specular color $\bc_s$ using the specular shading model given by \eq{specular}.

\paragraph{Volume rendering.}
Following~\cite{Chan2022eg3d, niemeyer2021giraffe}, we volume render the image $\bI_\bc$ and feature images $\bI_\bw$ by tracing over color $\bc$ and features $\bw$ respectively via \eq{volume_rendering}.

\paragraph{Superresolution.}
{%
Lastly, to generate high-resolution images, as in EG3D~\cite{Chan2022eg3d} we upsample the rendered images $\bI_\bc$ to $\bI_\bc^+$ via an upsampling module $\upsampleU$, guided by the feature images $\bI_\bw$.
We write
}%

\begin{equation}
    \bI_\bc^+ = 
    \upsampleU(\bI_\bc, \bI_\bw)
    .
\end{equation}

\subsection{Training}

{%
We extend the standard training process of EG3D~\cite{Chan2022eg3d} and adapt it to our framework.
Specifically, for the GAN setup, we use the rendered image $\bI_\bc$, and condition the discriminator on both the camera poses $\pose$ and the illumination SH coefficients $\ill$.
We further introduce a regularization on the estimated normal $\normal$, such that it matches the estimated densities.
}%
Similar to Zhang \etal~\cite{zhang2021nerfactor}, we introduce a loss term defined as
\begin{equation}
    \mathcal{L}_{\normal} = | \normal(\bx) - \nabla_\bx \density (\bx) |_1
    ,
\end{equation}
where $\nabla_{\bx}$ is the spatial gradient.

\section{Experiments}
\label{sec:experiments}

\subsection{Experimental setup}

\paragraph{Datasets.}
We use three datasets for our experiments: FFHQ~\cite{karras2019stylegan}, MetFaces~\cite{karras2020training}  and CelebA-HQ~\cite{karras2018progressive}.
FFHQ contains 70,000 samples and CelebA-HQ contains 30,000 samples of real human faces {as both datasets have been used traditionally to evaluate GAN methods}.
MetFaces contains 1,336 samples of faces taken from museum art images, {which is a small dataset that we use to demonstrate that our method can be applied beyond real face photos}.

\paragraph{Implementation details.}
As in Chan~\etal~\cite{Chan2022eg3d}, we mirror samples in each of the datasets to double the number of training samples.
We estimate the camera poses $\pose$ and illumination coefficients $\ill$ using DECA~\cite{feng2021deca}.

{We apply slightly varying training strategies for each dataset to account for their image resolution and volume.}
For FFHQ, we follow a strategy of EG3D~\cite{Chan2022eg3d} and train in two stages with a batch size of 32 on 8 GPUs.
In the first stage, we train for 750k iterations where
we volume render the images at $64^2$ and super resolve them to $512^2$.
In the second stage, we adjust the rendering resolution to $128^2$
{and super resolve them to $512^2$,} and train them further for 750k iterations.
For the CelebA-HQ dataset, we train only using the first stage at a rendering resolution of $64^2$ that are superresolved to $512^2$, for {500k iterations}.
{%
For the MetFaces dataset, as the sample size is small, we use a model pretrained on FFHQ and fine tune it on MetFaces via ADA augmentation~\cite{karras2020training}.
{We train for 15k iterations.}
}
We detail the network architectures in the supplementary material.

\begin{figure*}[t]
    \centering
    \includegraphics[width=0.64\textwidth]{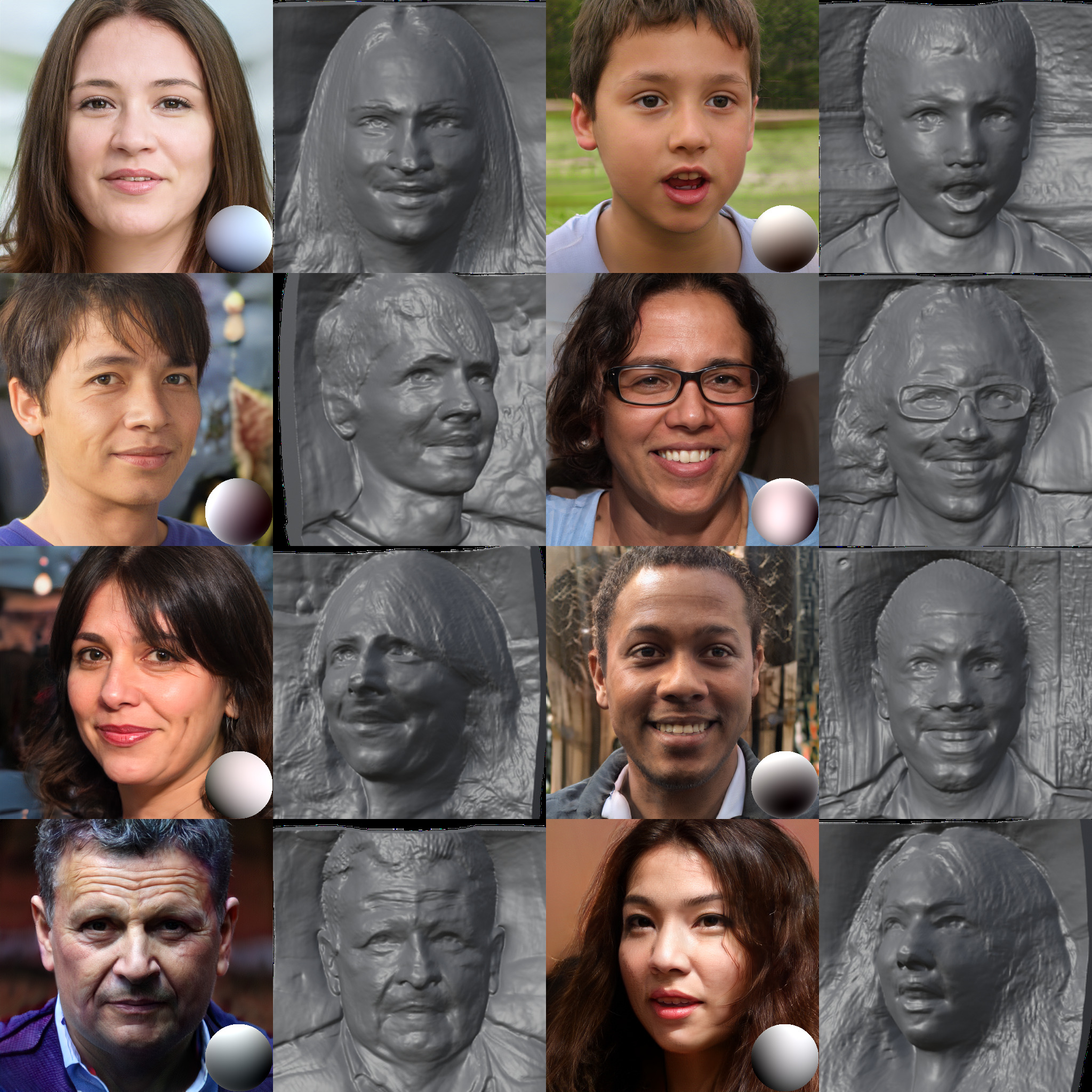} \hspace{0.05em}
    \includegraphics[width=0.32\textwidth]{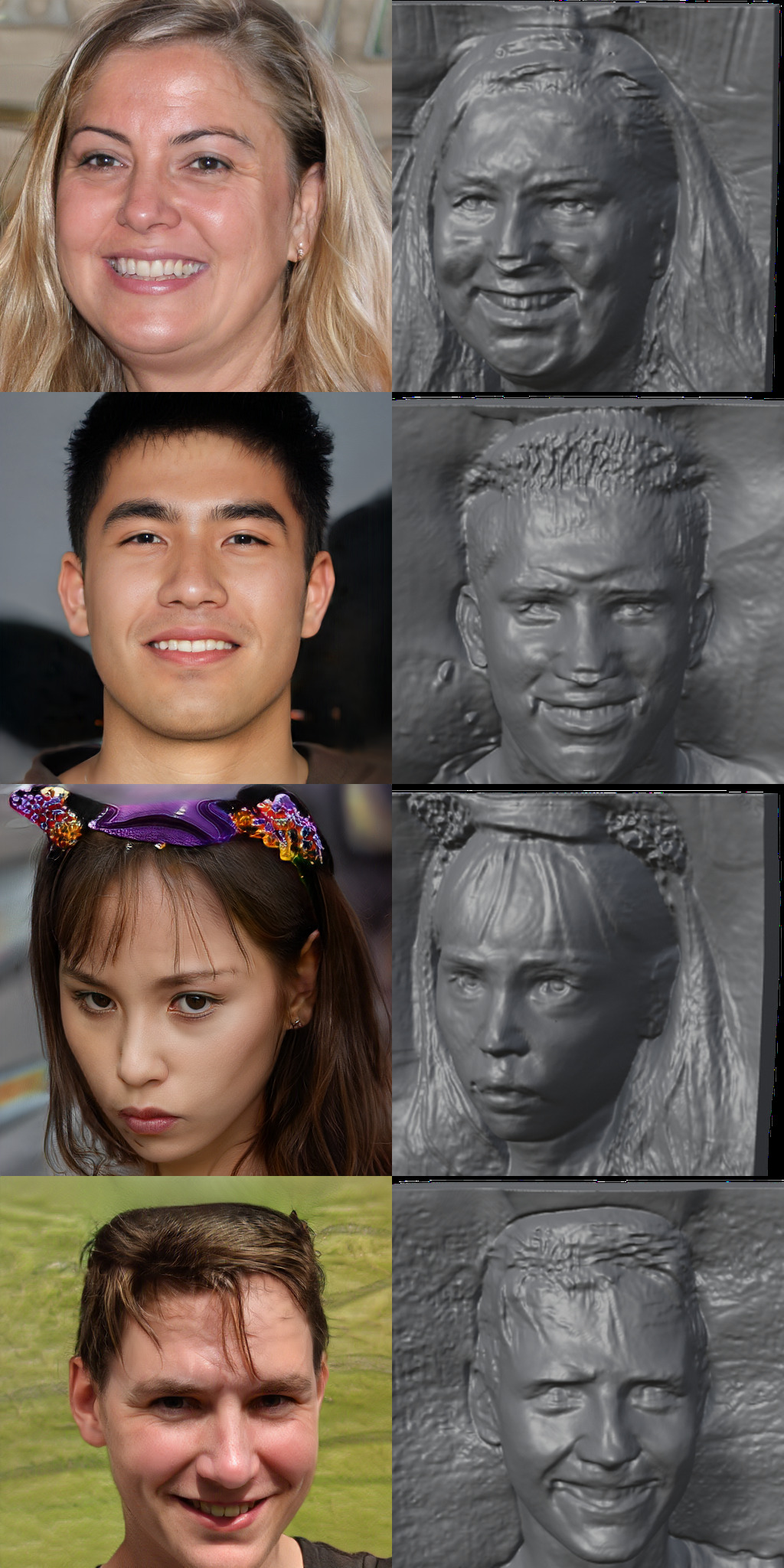}
    \begin{tabular}{p{0.64\textwidth}p{0.32\textwidth}}
    \centering
        \ours~(ours) & \centering  EG3D 
    \end{tabular}
    \caption{\textbf{Qualitative results.} Curated generated faces with different pose and illumination conditioning using our model (left) compared with the curated generated samples from EG3D (right). Our model shows detailed reconstruction in the lip and teeth region. The environment map is rendered using the half-sphere at the bottom right.
    }%
    \label{fig:ffhq_samples}
    \vspace{-0.6em}
\end{figure*}
\begin{figure*}[t]
    \centering
    \includegraphics[width=0.48\textwidth]{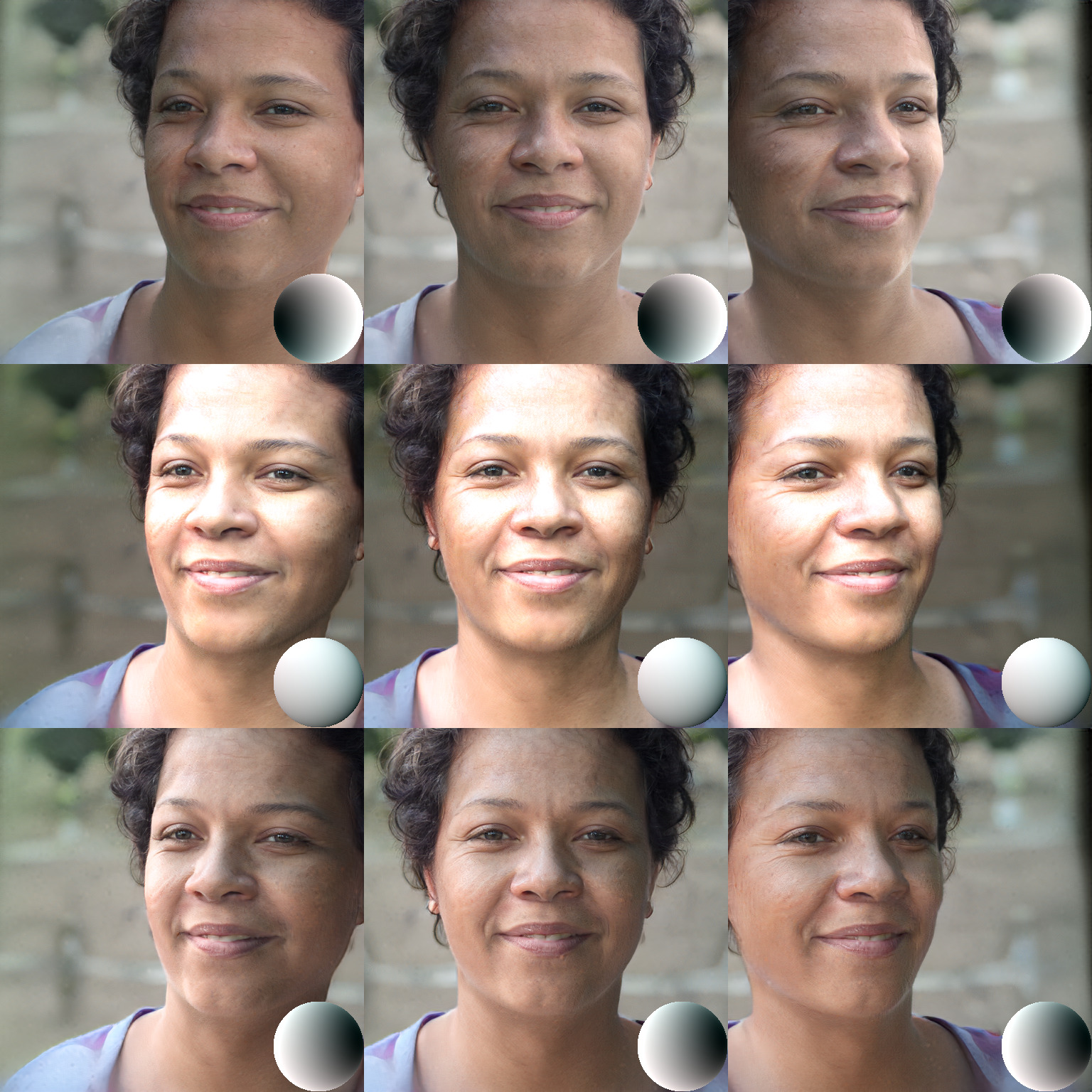}
    \hspace{0.05em}
    \includegraphics[width=0.48\textwidth]{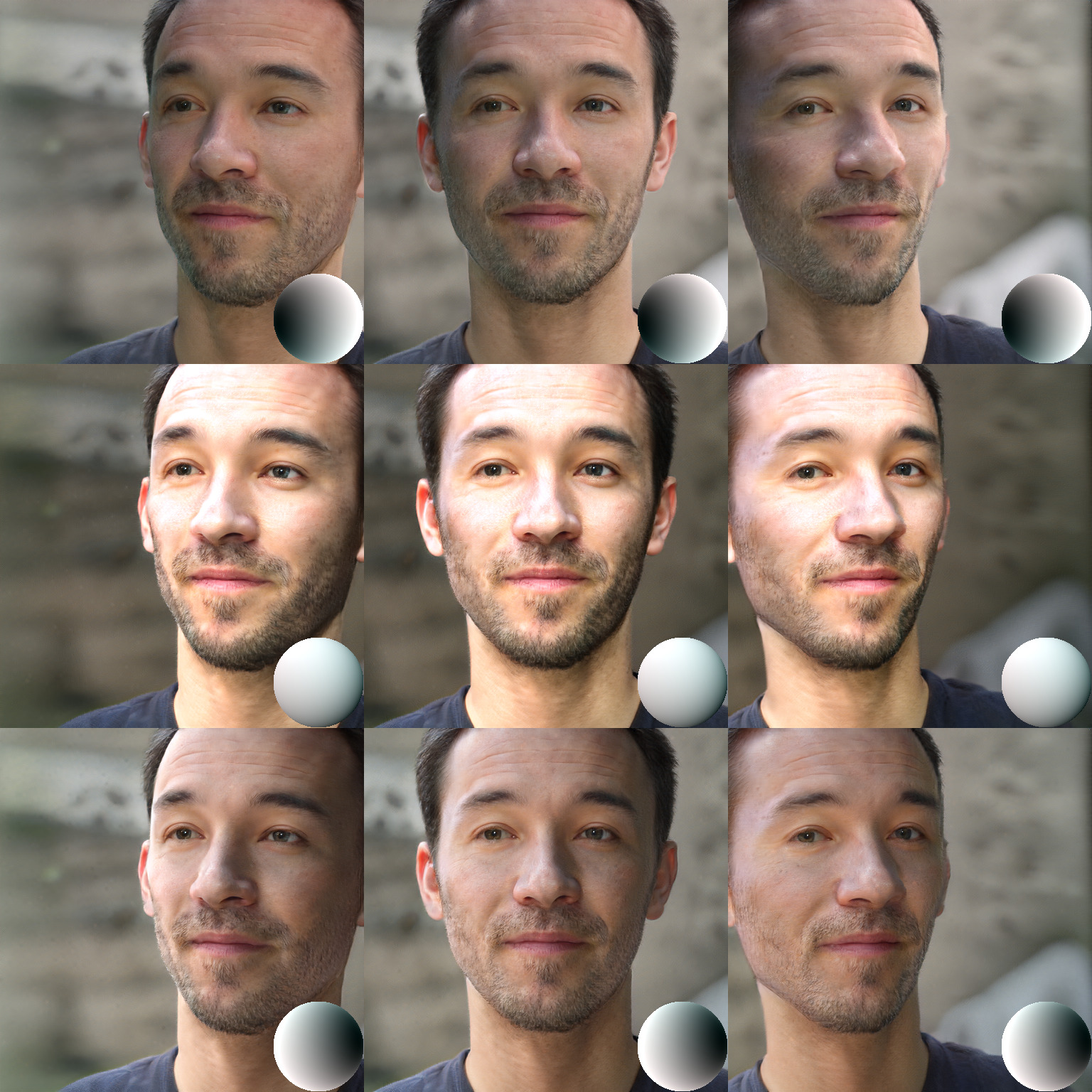}
    \caption{Camera Pose (left-to-right) and Illumination (top-to-bottom) interpolations. The illumination is represented by a normalized map~\cite{ramamoorthi2001efficient} at the bottom-right of each image. Each row has same illumination. Top and bottom row show directional light. Middle row shows overexposed central light.}
    \label{fig:pose_light_interp}
    \vspace{-1em}
\end{figure*}
\begin{figure}[t]
    \centering
    \includegraphics[width=0.48\textwidth]{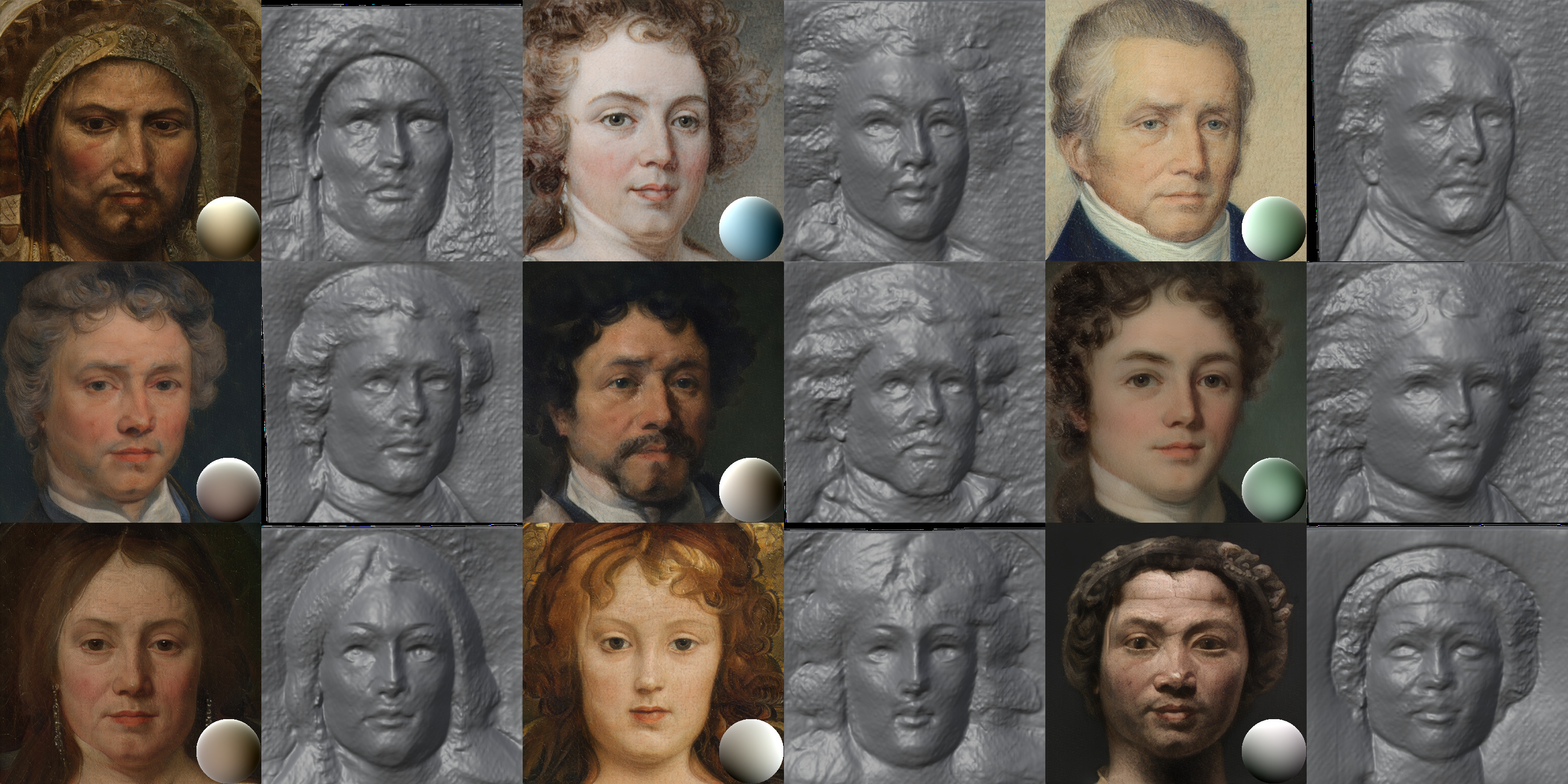}
    
    \vspace{0.2em}
    
    \includegraphics[width=0.48\textwidth]{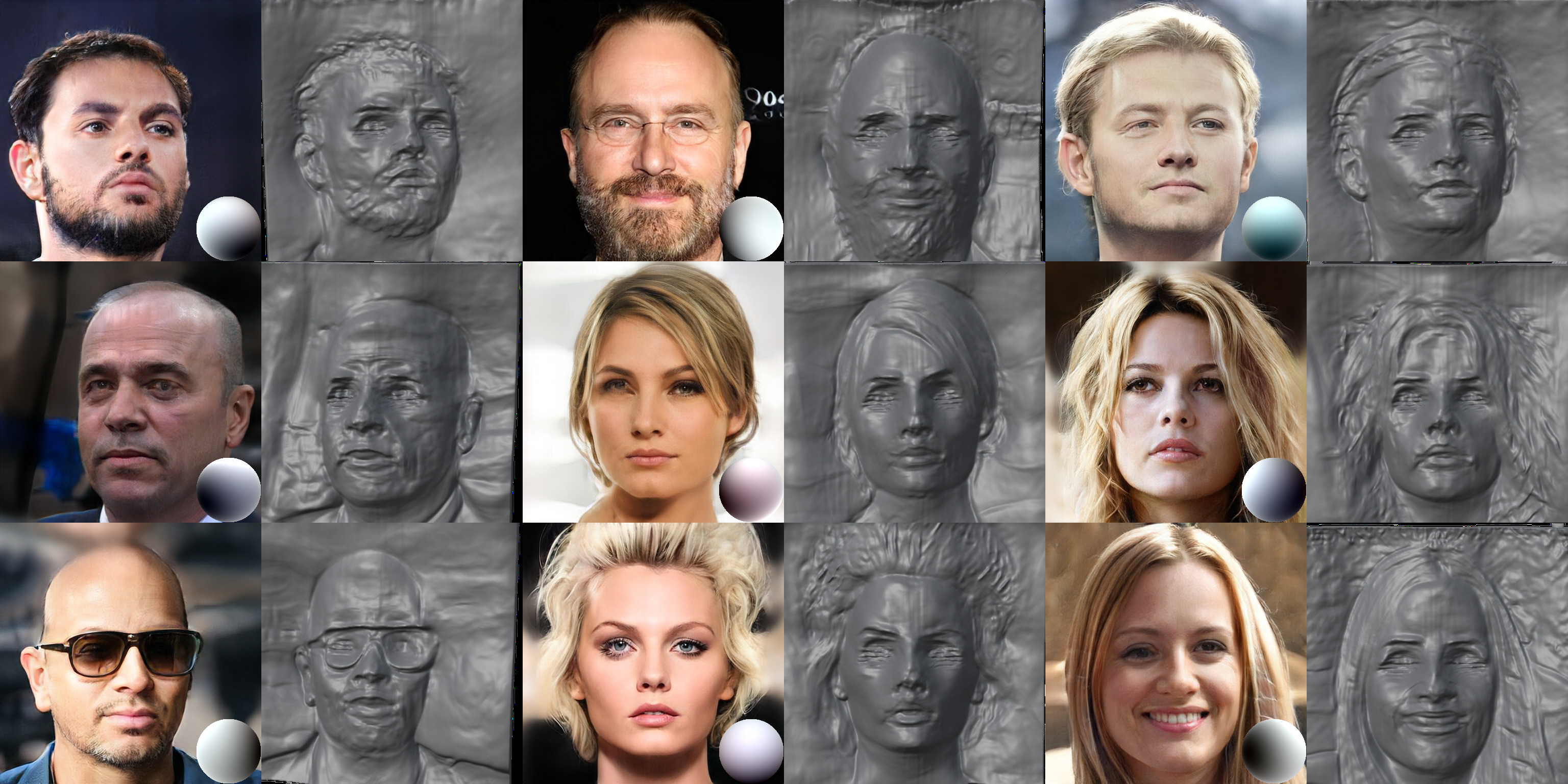}
    \caption{\textbf{Qualitative results.} Randomly generated faces with different pose and illumination conditioning using our model trained using MetFaces dataset (top) and CelebA-HQ dataset (bottom).
    }%
    \label{fig:metfaces_celeba_samples}
    \vspace{-0.6cm}
\end{figure}

\subsection{Qualitative results}

{%
We first qualitatively demonstrate the effectiveness of our method.
}%

\paragraph{Randomly drawn samples.}
We demonstrate the quality of our generated samples from the FFHQ dataset in \Figure{ffhq_samples}.
For our results, we also visualize the 3D reconstruction and the illumination used when generating these samples via a matt sphere for easy verification of the illumination consistency of each sample.
As shown, our results are photorealistic, also with fine details. Furthermore, the 2D images rendered from our model is visually consistant with the 3D shapes.
Note especially the regions around the lips and the teeth where our model provides improved 3D shape compared to EG3D, benefitting {from the specular modeling.}

We further show samples from CelebA-HQ and MetFaces datasets in \Figure{metfaces_celeba_samples}.
{%
For both datasets our results deliver rendering quality that photorealistic, or indistinguishable form actual paintings, while still providing explicit control over illumination and camera pose. We provide additional visualizations in the supplementary material.
}%

\paragraph{Controlling pose and illumination.}
In \Figure{pose_light_interp}, we fix the latent code for generating a face and vary the camera pose and illumination. Each row corresponds to a different illumination condition. We note that the lighting matches over two different persons. We also see the effect of strong lights in the middle row versus weak directional lights on the top and the bottom row.
Our model provides rendering that remains consistent regardless of the pose and illumination change.
We show additional results 
in the supplementary material.

\begin{figure}
    \centering
    \rotatebox{90}{Higher Specularity \qquad Normal \qquad Lower Specularity} \includegraphics[width=0.45\textwidth]{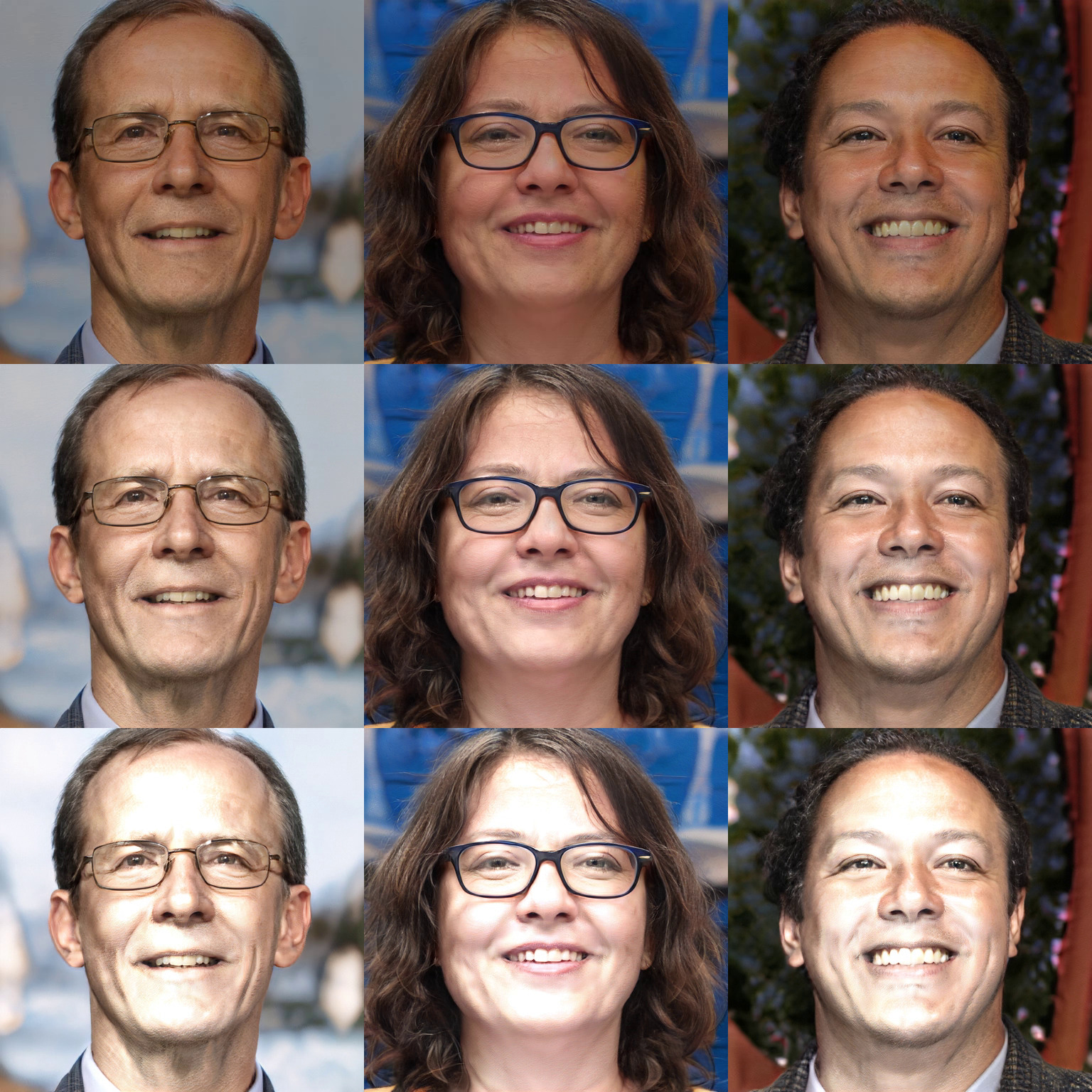}
    \caption{Effect of specular model by varying the specularity increasing from top to bottom. Specularity term helps in improving highlights on the nose and cheek region (middle row), and higher specularity results in the glaring effects (bottom row). }
    \label{fig:specular}
    \vspace{-0.6cm}
\end{figure}

\paragraph{Specular highlights.}
In \Figure{specular}, we show the effect of varying the specular component on the generation process. We vary the strength of specular component across each row in \Figure{specular}.
Note how the generated results for lower specularity seem matt, which is to be expected. Under normal conditions of specularity, highlights on the nose and cheeks are pronounced adding to the realism of the generated faces. Furthermore, using higher specularity results in glare.

\subsection{Quantitative results}
We further evaluate our results quantitatively through various metrics.

\paragraph{Evaluation metrics.}
We benchmark our generation quality on key metrics: FID score~\cite{heusel2017fid}, KID score~\cite{binkowski2018demystifying, heusel2017fid}, identity consistency (ID), depth, pose and light consistency.
\vspace{-0.7em}
\begin{itemize}[leftmargin=*]
\setlength\itemsep{-.3em}
    \item {\bf FID~\cite{heusel2017fid} and KID~\cite{binkowski2018demystifying, heusel2017fid}:} As in~\cite{Chan2022eg3d}, we sample 50,000 faces from the model and compute the score against the full dataset. 
    The FID and KID score captures the photorealism in the generated samples.
    \item {\bf Identity consistency (ID):}
    As in \cite{Chan2022eg3d, shi2021lifting, zhao2022gmpi, chan2021pigan}, we compute ID with the mean Arcface~\cite{deng2019arcface} cosine similarity, after rendering a face with two random camera views.
    The measure highlights the consistency of the face under different rotations.
    
    \item {\bf Consistency:}
    We further evaluate our model on camera pose, depth and illumination consistency.
    We sample 1024 faces from the generator along with their camera pose, illumination and depth using the our model.
    We then use DECA~\cite{feng2021deca} to obtain psuedo ground truths for camera pose, illumination of the generated samples. {This differs from baseline methods~\cite{Chan2022eg3d,zhao2022gmpi, chan2021pigan, shi2021lifting} that use Deng et al.~\cite{deng2019facerecon} for pose estimation and preprocessing. Since DECA has better performance, this also contributes to lower error in our evaluation.} For depth consistency, following  previous work~\cite{Chan2022eg3d, zhao2022gmpi, shi2021lifting}, we estimate pseudo ground truth depth from Deng et al.~\cite{deng2019facerecon}. We report the mean square error between our estimates and the pseudo ground truth estimates.
\end{itemize}

\vspace{-0.6em}
\paragraph{Variants.}
{%
For the quantitative study we evaluate four different variants of our method.
}%
We introduce two sets of models \ours-d and \ours-f using the diffuse-only model and the full model respectively with volume rendering resolution of $64^2$ and superresolved to $512^2$.
We further train the models \ours-D and \ours-F at volume rendering resolution of $128^2$ and superresolved to $512^2$.

\paragraph{FFHQ.}
We report the quatitative evaluations on the FFHQ dataset in~\Table{ffhq_eval} and compare with previous work on 3D aware GANs.
We observe that \ours-D and \ours-F obtain state-of-the-art in photorealism metrics---FID, KID and competitive performance on ID and depth consistency metrics. Amongst methods that generate at $512^2$ resolution, we achieve state-of-the-art accuracy on depth consistency metrics. We also note that although diffuse models achieve state-of-the-art on photorealism metrics, the full models that also model the specular components have better depth.
On the pose consistency metrics, we achieve state-of-the-art performance.

\begin{table}[]
    \centering
    \small
    \scalebox{0.85}{
    \begin{tabular}{@{}lccccc@{}}
    \toprule
%         &  \multicolumn{5}{c}{FFHQ} & Cats \\
           &  FID $\downarrow$ & KID $\downarrow$ & ID $\uparrow$ & Depth $\downarrow$ & Pose $\downarrow$   \\
    \midrule
    GIRAFFE~\cite{niemeyer2021giraffe} $256^2$ & 31.5 & 1.992 & 0.64 & 0.94 & 0.089   \\
    $\pi$-GAN~\cite{chan2021pigan} $128^2$ & 29.9 & 3.573 & 0.67 & 0.44 & 0.021  \\ 
    Lift. SG~\cite{shi2021lifting} $256^2$ & 29.8 & - & 0.58 & 0.40 & 0.023  \\ 
    StyleNeRF~\cite{gu2021stylenerf} $512^2$ & 7.80 & 0.220 & - & - & - \\ 
    StyleNeRF~\cite{gu2021stylenerf} $1024^2$ & 8.10 & 0.240 & - & - & - \\ 
    GMPI~\cite{zhao2022gmpi} $512^2$ & 8.29 & 0.454 & 0.74 & 0.46 & 0.006 \\
    GMPI~\cite{zhao2022gmpi} $1024^2$ & 7.50 & 0.407 & 0.75 & 0.54 & 0.007 \\
    EG3D~\cite{Chan2022eg3d} $256^2$ & 4.80 & 0.149 & 0.76 &\bf 0.31 & 0.005 \\
    EG3D~\cite{Chan2022eg3d} $512^2$ & 4.70 & 0.132 &\bf 0.77 & 0.39 & 0.005 \\
    \midrule
    \ours-d $512^2$ & 4.01 & 0.124 & 0.72 & 0.42 & 0.0009\\ 
    \ours-f $512^2$ & 4.06 & 0.115 & 0.72 & 0.33 &\bf 0.0008 \\
    \ours-D $512^2$ & \textbf{3.48} & \bf 0.097 &\bf 0.77 & 0.62 &\bf 0.0008 \\
    \ours-F $512^2$ & 3.90 & 0.117 & 0.75 & 0.43 &\bf 0.0008 \\
    \bottomrule
    \end{tabular}}
    \caption{Comparison of 3D aware GANs with \ours~(ours) using FID, KID $\times 100$, Identity Consistency, depth accuracy and pose accuracy on the FFHQ dataset~\cite{karras2019stylegan}. The method names are suffixed with the resolution of generated images. }
    \label{tab:ffhq_eval}
\end{table}

\paragraph{MetFaces.}
In \Table{metfaces_eval}, we compare the performance of our model to a 2D GAN --  StyleGAN2~\cite{karras2020stylegan2}, and a 3D aware GAN, StyleNeRF~\cite{gu2021stylenerf}. We obtain better photorealism than both the methods on FID and KID scores. We further provide face consistency metrics and 3D consistency metrics -- depth and pose. We note that the depth metrics obtained on this dataset are worse than those of FFHQ. A reason for this could be that art images do not strictly respect the physical model of illumination.

\begin{table}[]
    \centering
    \small
    \scalebox{0.92}{
    \begin{tabular}{@{}lccccc@{}}
    \toprule
           &  FID $\downarrow$ & KID $\downarrow$ & ID $\uparrow$ & Depth $\downarrow$ & Pose $\downarrow$     \\
    \midrule
    StyleGAN2~\cite{karras2020stylegan2} & 18.9 & 0.27 & - & - & - \\
    StyleNeRF~\cite{gu2021stylenerf} & 20.4 & 0.33 & - & - & -\\
    \ours-d  & \bf 15.30 & \bf	0.22 &\bf 0.87 & 0.96 &\bf 0.0018 \\
    \ours-f & 15.43 & 0.23 & 0.86 &\bf 0.77 & 0.0032\\
    \bottomrule
    \end{tabular}}
    \caption{Comparison of a 2D GAN and a 3D aware GAN with \ours~(ours) using FID, KID $\times$ 100, Identity Consistency, depth accuracy and pose accuracy on the MetFaces dataset~\cite{karras2020training}. All models generate at a resolution of $512^2$.}
    \label{tab:metfaces_eval}
\end{table}

\begin{table}[]
    \centering
    \small
    \scalebox{0.85}{
    \begin{tabular}{@{}lccccc@{}}
    \toprule
           &  FID $\downarrow$ & KID $\downarrow$ & ID $\uparrow$ & Depth $\downarrow$ & Pose $\downarrow$  \\
    \midrule
    StyleGAN~\cite{karras2019stylegan} $1024^2$ & 4.41 & - & - & - & - \\
    \ours-d $512^2$ &\bf {3.63} &\bf 0.083 &\bf 0.73 & 0.36 &\bf 0.0011 \\
    \ours-f $512^2$ & 3.94 & 0.117 & 0.72 &\bf 0.33 &\bf 0.0011 \\
    \bottomrule
    \end{tabular}}
    \caption{Performance of \ours~evaluated using FID, KID $\times 100$, Identity Consistency, depth accuracy and pose accuracy on the CelebA-HQ dataset~\cite{karras2018progressive}}
    \label{tab:celeba_eval}
    \vspace{-1em}
\end{table}

\paragraph{CelebA-HQ.}
In \Table{celeba_eval}, we evaluate our models trained on a smaller dataset, CelebA-HQ with 30,000 samples. We note that our models can achieve better photorealism scores than 2D GANs such as StyleGAN and has good performance on consistency metrics. Furthermore, we also note that the full model,~\ours-f provides better depth accuracy than diffuse only,~\ours-d model.

\subsection{Illumination accuracy}
\label{sec:ablation}

\begin{figure}[t]
    \centering
    \includegraphics[width=0.45\textwidth]{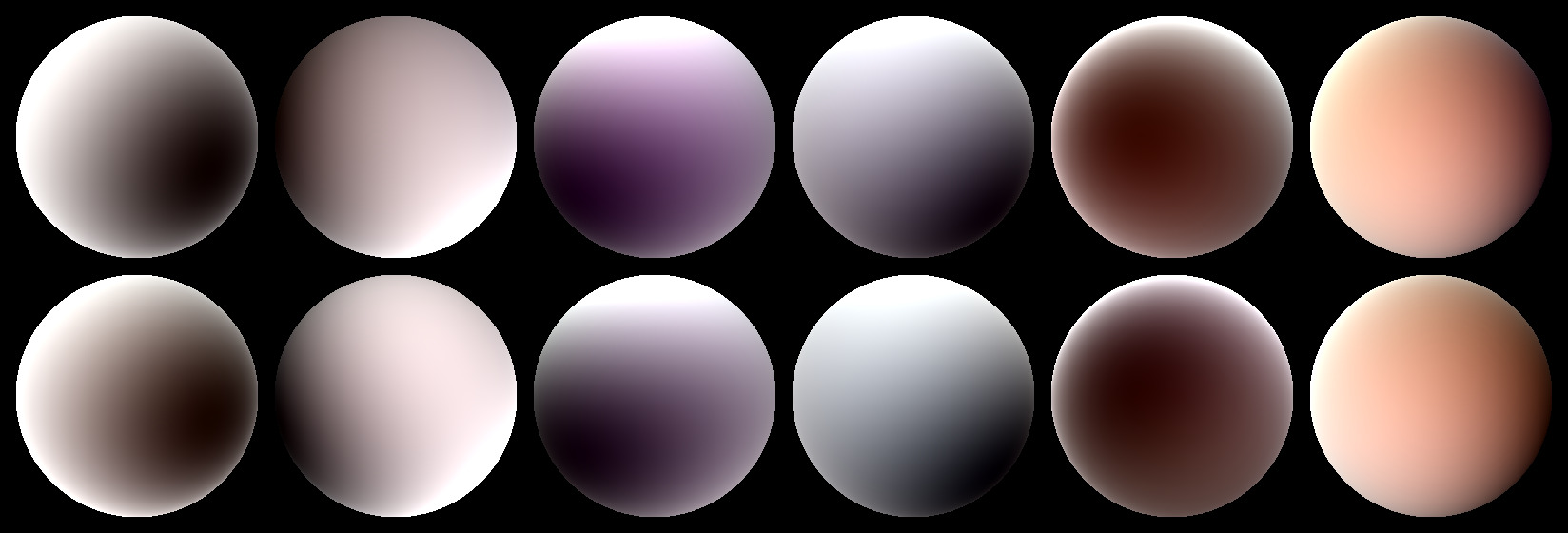}
\caption{Top: Irradiance maps used to condition our generator. Bottom: Pseudo ground-truth estimates of irradiance maps using DECA~\cite{feng2021deca} on the generated images.}
    \label{fig:illumination_visuals}
    \vspace{-0.4cm}
\end{figure}

{%
We further report how accurate our model learns the illumination effects by generating random samples and running DECA~\cite{feng2021deca} on them to see how well the estimated illumination agree with the conditioned illumination. We use mean square error of the SH coefficients averaged over 1024 random samples from our model. The samples are conditioned on pose and illumination randomly sampled from the training dataset.
While imperfect, as these results will be limited by the accuracy of DECA~\cite{feng2021deca}, it allows us to roughly gauge the {accuracy of our illuminations}.
We report these results in \Table{light_accuracy}.
}%
{%
As reported, our model generates images that are well inline with the DECA estimates.
}%
We show the visualizations in \Figure{illumination_visuals}. 
\begin{table}[]
    \small
    \centering
    \begin{tabular}{@{}lccc@{}}
    \toprule
    & FFHQ     & CelebA-HQ & MetFaces  \\
    \midrule 
    \ours-d   & 0.0054 & 0.0061 & 0.0069 \\
    \ours-f   & 0.0053 & 0.0042  & 0.0084\\
    \ours-D & 0.0049 & - & - \\
    \ours-F & 0.0051  & - & -\\
    \bottomrule
    \end{tabular}
    \caption{Average mean square error (MSE) of illumination SH coeffients evaluated using pseudo ground-truth from DECA~\cite{feng2021deca}}
    \label{tab:light_accuracy}
    \vspace{-0.4cm}
\end{table}

\section{Conclusion}

{%
We have presented a novel method to learn a disentangled 3D generative model for faces that allows the user to control the camera pose and the illumination, with only single views, and without the need of any manual annotation.
Our core idea is how we model {physics-based} rendering with a simplified Phong model, that integrates effectively to neural volume rendering. 
We demonstrate the effectiveness of our method on FFHQ, CelebA-HQ, and MetFaces dataset, providing photorealistic image quality.
}%

\paragraph{Limitations and future work.}
{Our method does not model all the physical aspects of the scene and rendering process.
Our setup is unsupervised, learned without any explicit control of the environment during the capture.
While this problem setup is difficult, we are capable of generating photorealistic faces under various diffuse/specular lighting.
The quality of the method may be further improved by modeling global illumination or subsurface scattering. 
For further improved factorization/specular quality, we may need higher frequency environment maps~\cite{han2007frequency} that are difficult to obtain in an unsupervised setting where manual annotation or specially designed capture setup would help.}
Further, our method uses estimation of illumination parameters and camera pose from existing methods~\cite{feng2021deca}.
As such, the accuracy of our method is limited by the performance of existing work. 
In \Figure{ffhq_samples}, it can be seen that the estimated illumination can sometimes be entangled with skin color, which is a limitation that we inherit from DECA~\cite{feng2021deca}.

\paragraph{Ethical Considerations.} We intend our work to be used for research purposes only and should not be used edit images of real people without their consent. The use of our method for developing applications should carefully consider privacy of the faces, as well as, bias present in the datasets.

%%%%%%%%% REFERENCES
{\small
\bibliographystyle{ieee_fullname}
\bibliography{egbib}
}

%\clearpage

\section*{A. Network Architecture}
\label{sec:architecture}

\begin{figure}
    \centering
    \includegraphics[width=0.12\textwidth]{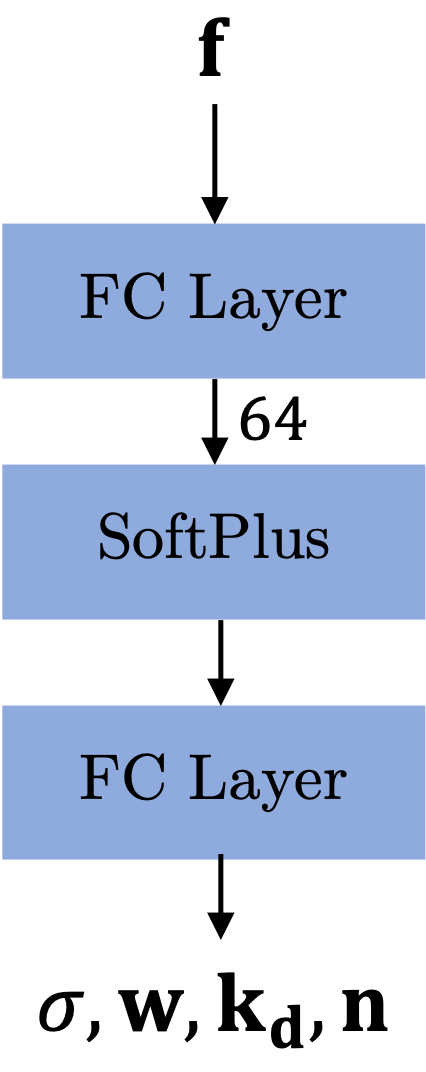} \qquad \qquad
    \includegraphics[width=0.12\textwidth]{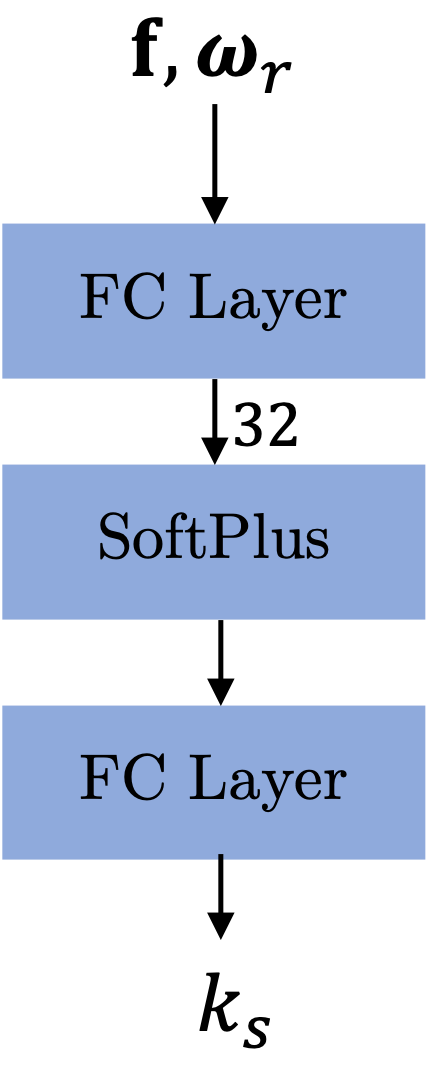} \\
    \begin{tabular}{cc}
        (a) Diffuse Decoder \quad & \quad (b) Specular Decoder
    \end{tabular}
    \caption{Architecture of Diffuse Decoder and Specular Decoder }
    \label{fig:decoders}
\end{figure}

We show the architecture of the diffuse Decoder and the specular decoder as described in Section 3 in Fig.~\ref{fig:decoders}. We follow a feed-forward architecture with fully connected layers and softplus activation~\cite{zhao2018softplus}. The fully connected layers have a hidden dimension of 64 for diffuse decoder and 32 for specular decoder.

\section*{B. Training Time}
Our training time is slightly higher than EG3D due to addition of illumination model. EG3D takes 4 days, FaceLit-d takes 4.5 days and FaceLit-f takes 4.8 days on 8 NVIDIA A100 GPUs to train at a neural rendering resolution of $64^3$. 

\section*{C. Additional Qualitative Results}

\begin{figure*}
    \centering
    \includegraphics[width=\textwidth]{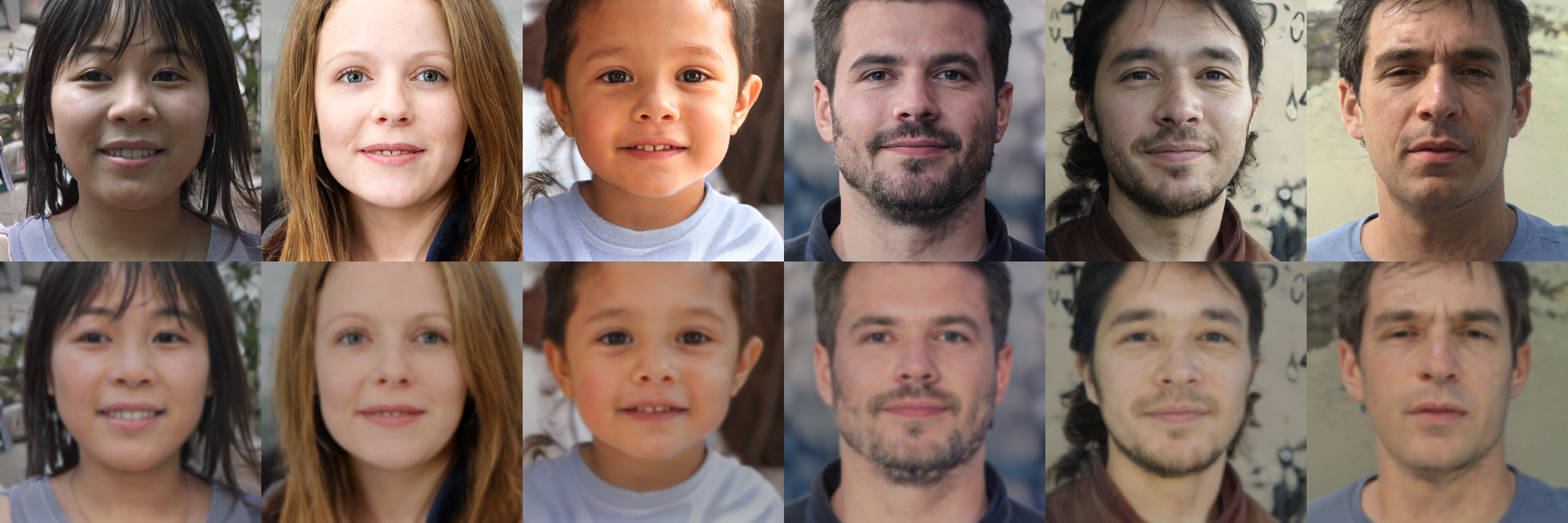}
    \caption{\textbf{Albedo Visualization:} Generated images on top with their corresponding albedo images at the bottom.}
    \label{fig:albedo}
\end{figure*}

\paragraph{Albedo.} In Fig.~\ref{fig:albedo}, we show the visualization of albedo images of the corresponding generated samples. The albedo images are obtained by rendering the images before the superresolution module under constant illumination.

\begin{figure*}
    \centering
    \includegraphics[width=\textwidth]{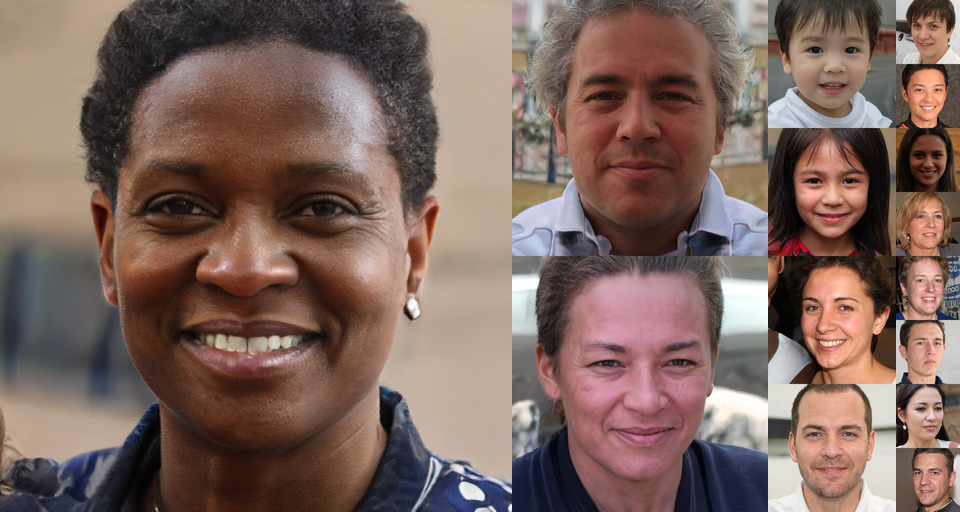}
    \caption{Uncurated FFHQ samples at $\psi = 1.0$}
    \label{fig:uncurated_ffhq_psi_1}
\end{figure*}

\begin{figure*}
    \centering
    \includegraphics[width=\textwidth]{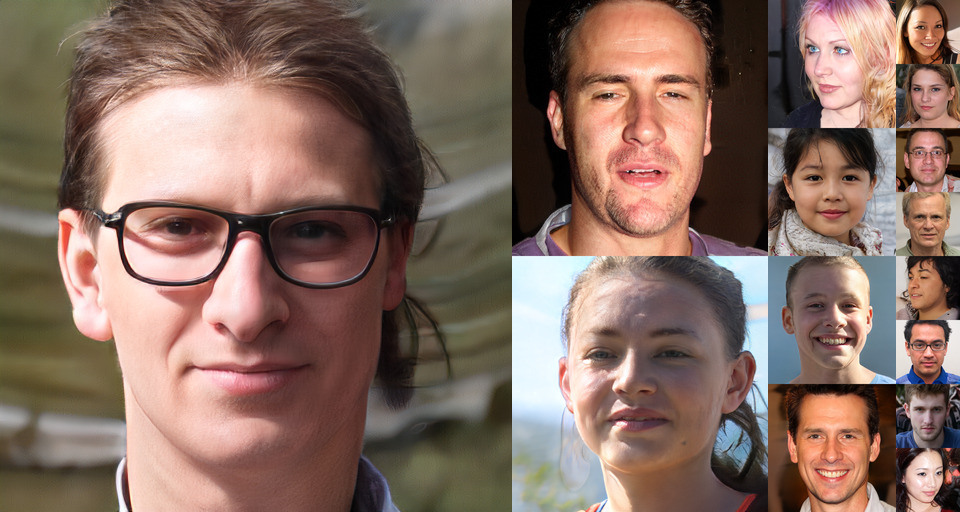}
    \caption{Uncurated FFHQ samples at $\psi = 0.7$}
    \label{fig:uncurated_ffhq_psi_07}
\end{figure*}

\paragraph{Uncurated Samples.} In Fig.~\ref{fig:uncurated_ffhq_psi_1}, we show uncurated samples generated by our model with a truncation factor of $\psi = 1.0$. In Fig.~\ref{fig:uncurated_ffhq_psi_07}, we show uncurated samples generated by our model with a truncation factor of $\psi = 0.7$. In both case, we use the FaceLit-F model trained on the FFHQ dataset.

\begin{figure*}
    \centering
    \includegraphics[width=\textwidth]{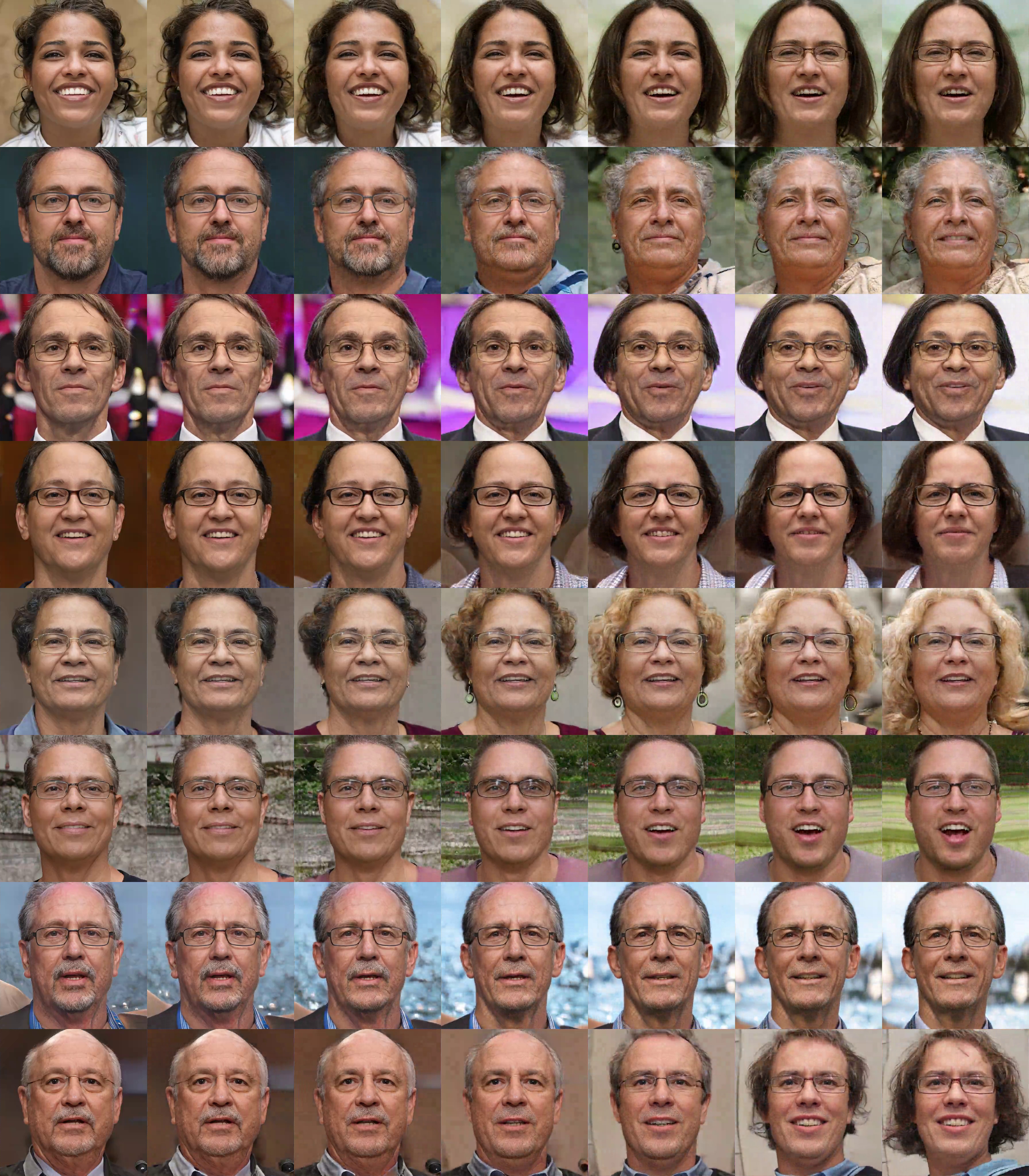}
    \caption{\textbf{Latent Space Interpolation:} In each row, we smoothly vary the latent code from left to right.}
    \label{fig:latent_interp}
\end{figure*}

\paragraph{Latent Space Interpolations.} In Fig.~\ref{fig:latent_interp}, we show interpolation across generated samples from our model. In each row, we interpolate the latent code linearly and observe that the generated samples vary smoothly and generalize to wide variety of faces, even with accessories such as eyeglasses.

\paragraph{Videos.} For visuals, please open the \texttt{README.html}. The webpage contains all the video results. The webpage is tested to run on a MacOS and Ubuntu and supports Safari, Chrome, and Firefox browsers. All videos are encoded with \texttt{h264} encoding and the raw files are present in the \texttt{data} directory. 

\section*{D. Additional Quantitative Results}
We compare our method with VoLux-GAN~\cite{tan2022volux} and ShadeGAN~\cite{pan2021shadegan} by evaluating the face identity consistency of the generated samples by measuring the cosine similarity between the samples using Arcface~\cite{deng2019arcface}. In Tab.~\ref{tab:voluxgan_id}, we compute the cosine similarity of the face rendered with yaw $\in [-0.5, -0.25, 0.25, 0.5]$ radians with yaw $ = 0$. The methods are trained on different datasets therefore it does not reflect a direct comparison. However, it shows that our method preserves identity similar to the baseline methods.
 
In Tab.~\ref{tab:voluxgan_ill_id}, we evaluate the identity similarity under different lighting conditions as compared to the rendered albedo of the face. Similar to previous evaluation, the methods are trained on different datasets and have different models of illumination. ShadeGAN uses point light sources, and the face consistency is measured under different point light sources and the albedo. In VoLux-GAN, the face consistency is measured under different environment maps and the albedo. In our case, we measure it using 3 different spherical harmonics lights and the albedo. We observe that the face similarity is preserved under different lighting conditions.

\begin{table*}
    \centering
    \scalebox{1.}{
    \begin{tabular}{lccccc}
    \toprule
     & Dataset & \multicolumn{4}{c}{Yaw changes in radians} \\
    & &  -0.5 & -0.25 & 0.25 & 0.5 \\ 
    \midrule
    ShadeGAN~\cite{pan2021shadegan}  & CelebA & 0.481 & 0.751 & 0.763 & 0.500 \\
    VoLux-GAN~\cite{tan2022volux}  & CelebA & 0.606 & 0.774 & {0.800} & 0.599 \\
    FaceLit-D  & FFHQ & {0.615} & {0.817} & 0.781 & 0.596 \\
    FaceLit-F  & FFHQ & 0.581 & 0.795 & 0.783 & {0.626} \\
    \bottomrule
    \end{tabular}}
    \caption{Face consistency metrics with yaw changes.}
    \label{tab:voluxgan_id}
\end{table*}

\begin{table*}
    \centering
    \scalebox{1.}{
    \begin{tabular}{lcccccc}
    \toprule
     & Dataset & Illumination Type & \multicolumn{3}{c}{Setting} \\
    & & & 1 & 2 & 3\\ 
    \midrule
    ShadeGAN~\cite{pan2021shadegan}  &  CelebA & Point Source & 0.581 & 0.649 & 0.666 \\
    VoLux-GAN~\cite{tan2022volux}  & CelebA & Environment Map &  0.760 & {0.890} & 0.808\\
    FaceLit-D  & FFHQ & Spherical Harmonics & 0.837 & {0.911} & 0.811   \\
    FaceLit-F  & FFHQ & Spherical Harmonics & {0.839}  & 0.906  & {0.815} \\
    \bottomrule
    \end{tabular}}
    \caption{Face consistency metrics with illumination changes.}
    \label{tab:voluxgan_ill_id}
\end{table*}

\end{document}